\documentclass{article}

    \usepackage[preprint]{neurips_2021}

\usepackage[utf8]{inputenc} 
\usepackage[T1]{fontenc}    
\usepackage{hyperref}       
\usepackage{url}            
\usepackage{booktabs}       
\usepackage{amsfonts}       
\usepackage{nicefrac}       
\usepackage{microtype}      
\usepackage{xcolor}         
\usepackage{graphicx}
\usepackage{float}
\usepackage{subfig}
\usepackage{adjustbox}
\usepackage{multirow}
\usepackage{caption}
\title{On the Susceptibility and Robustness of Time Series Models through Adversarial Attack and Defense}

\author{%
  Asadullah Hill Galib\\
  Computer Science and Engineering\\
  Michigan State University\\
  \texttt{galibasa@msu.edu} \\ 
   \And
   Bidhan Bashyal \\
    Computer Science and Engineering\\
   Michigan State University\\
   \texttt{bashyalb@msu.edu} \\
}

\begin{document}

\maketitle

\begin{abstract}
Under adversarial attacks, time series regression and classification are vulnerable. Adversarial defense, on the other hand, can make the models more resilient. It is important to evaluate how vulnerable different time series models are to attacks and how well they recover using defense. The sensitivity to various attacks and the robustness using the defense of several time series models are investigated in this study. Experiments are run on seven time series models with three adversarial attacks and one adversarial defense. According to the findings, all models, particularly GRU and RNN, appear to be vulnerable. LSTM and GRU also have better defense recovery. FGSM exceeds the competitors in terms of attacks. PGD attacks are more difficult to recover from than other sorts of attacks.
\end{abstract}

\section{Introduction}

Time series analysis is a vital problem in data mining with many applications including finance, weather forecasting, industrial maintenance, and many others. Deep Learning (DL) methods have shown great success in analyzing time series data \cite{financial}. One downside of DL methods is that they can be fooled easily by generating adversarial attacks. Adversarial attacks in deep learning have been vastly studied for image recognition and classification problem. Similarly, it is very important to explore adversarial attacks in the time-series analysis as time series analysis uses various DL algorithms. Designing a defense mechanism against these attacks is even essential for robustness. From an adversarial standpoint, there has not been much research into DL models for time series regression and classification.

DL models for time series are also susceptible under adversarial attacks \cite{fawaz2019adversarial, harford2020adversarial, mode2020adversarial, siddiqui2020benchmarking, wu2022small, rathore2020untargeted}. Most of this research concentrate on performing adversarial attacks for either regression or classification problems. Defense mechanism against the adversarial attacks for time series is also explored \cite{siddiqui2020benchmarking}. Mostly, this research focuses on different adversarial attacks and defenses rather than the model itself. But it's also worth considering how vulnerable different time series models are to attacks and how effectively they can recover utilizing defense. This research examines the susceptibility and robustness of various time series models to various attacks and defenses. To the best of our knowledge, no previous research has explored this. Our contribution and the things we are addressing are summarized in Table \ref{contribution}.

\begin{table*}[h]
\centering
\begin{adjustbox}{width=1.0\textwidth, height=1.1cm}
\begin{tabular}{c|c|c|cclccllc|cccc|c}
\hline
\multirow{2}{*}{\textbf{References}} & \multirow{2}{*}{\textbf{Regression}} & \multirow{2}{*}{\textbf{Classification}} & \multicolumn{6}{c}{\textbf{Models}}                                                                & \textbf{}       & \textbf{}      & \multicolumn{4}{c|}{\textbf{Adversarial Attack}}             & \multirow{2}{*}{\begin{tabular}[c]{@{}l@{}}\textbf{Adversarial} \\ \textbf{Defense}\end{tabular}} \\ \cline{4-15}
                                     &                                      &                                          & \textbf{LSTM} & \textbf{GRU} & \textbf{RNN} & \textbf{CNN} & \textbf{CNN-LSTM} & \textbf{ConvLSTM} & \textbf{ResNet} & \textbf{Other} & \textbf{FGSM} & \textbf{BIM} & \textbf{PGD} & \textbf{Other} &                                                \\ \hline
Fawaz et al., 2019 \cite{fawaz2019adversarial}                                    &                                      & \checkmark                                        &               &              &              &              &                   &                   & \checkmark               &                & \checkmark             & \checkmark            &              &                &                                                \\  
Harford et al., 2020 \cite{harford2020adversarial}                                    &                                      & \checkmark                                        &               &              &              &              &                   &                   &                 & \checkmark              &               &              &              & \checkmark              &                                                \\  
Mode et al., 2020 \cite{mode2020adversarial}                                    & \checkmark                                    &                                          & \checkmark             & \checkmark            &              & \checkmark            &                   &                   &                 &                & \checkmark             & \checkmark            &              &                &                                                \\ 
Siddiqui et al., 2020 \cite{siddiqui2020benchmarking}                                    &                                      & \checkmark                                        &               &              &              &              &                   &                   &                 & \checkmark              & \checkmark             &              & \checkmark            & \checkmark              & \checkmark                                              \\  
Wu et al., 2022\cite{wu2022small}                                     & \checkmark                                    &                                          & \checkmark             &              & \checkmark            & \checkmark            &                   &                   &                 &                & \checkmark             &              &              & \checkmark              &                                                \\  
Rathore et al., 2020 \cite{rathore2020untargeted}                                   &                                      & \checkmark                                        &               &              &              &              &                   &                   & \checkmark               &                & \checkmark             & \checkmark            &              &                &                                                \\ \hline
\textbf{Our Study}                                    & \checkmark                                    & \checkmark                                        & \checkmark             & \checkmark            & \checkmark            & \checkmark            & \checkmark                 &      \checkmark             &                 &                & \checkmark             & \checkmark            & \checkmark            &                & \checkmark                                              \\ \hline
\end{tabular}
\end{adjustbox}
\caption{Contribution of this Study}
\label{contribution}
\end{table*}

In this paper, we use 7 different DL time series models, such as Long -Short Term Memory(LSTM), Stacked-LSTM (LSTM with more than one layer), Gated Recurrent Unit (GRU), Recurrent Neural Network (RNN), Convolutional Neural Network (CNN), CNN-LSTM and ConvLSTM to train for regression and classification problems for time series analysis. Different adversarial mechanisms such as Fast Gradient Sign Method (FGSM), Basic Iterative Method (BIM), and Projected Gradient Descent (PGD) are used to attack the models. Additionally, we also implemented adversarial training as a defense mechanism to tackle the attacks. We evaluate the performance of each adversarial attack over the 7 different models in 6 different datasets (5 regression and 1 classification). Results suggest all models are vulnerable, especially GRU and RNN. Also, LSTM and GRU show better recovery through the defense. FGSM outperforms others in terms of attacks. Recovering from a PGD attack is harder than others. 

The rest of the paper is organized as follows. Section 2 briefly discusses the related works, Section 3 introduces the problem statement. Section 4 covers in detail the models we used for training, adversarial attacks, and defense mechanism. Section 5 describes data sets, experimental setup, and results. Section 6 provides discussions on the results. Section 7 briefly provides the individual contribution. Section 8 concludes this paper.

\section{Related Works}

Several studies investigate adversarial attacks and defenses for time series problems. \cite{fawaz2019adversarial} proposes to use adversarial attack techniques such as (FGSM and BIM) to reduce model performance when classifying instances at test time. As a time series classifier, it employs ResNet. The classifier is subject to adversarial attacks, according to its findings. It does not take into account PGD attacks and only attacks one classifier model. It also focuses solely on time series classification rather than regression. \cite{harford2020adversarial} studies black-box and white-box attacks
on multivariate time series.  Adversarial Transformation Network (ATN) and Gradient Adversarial Transformation Network (GATN) are used to create adversarial samples. It makes use of 18 datasets to attack 1-Nearest Neighbor Dynamic Time Warping (1-NN DTW) and a Fully Convolutional Network (FCN). On all 18 datasets, both models were vulnerable to attack. It does not consider time series regression and typical attacks (FGSM, BIM, PGD). Using time-series data, \cite{siddiqui2020benchmarking} performs extensive benchmarking of well-proven adversarial defensive approaches. It uses 2 white-box attacks (FGSM and PGD) and 3 black-box attacks (Noise Attack, Boundary Attack, and Simple Black-box Attack (SIMB). It evaluates robustness using multiple adversarial defenses (adversarial training, TRADES, Feature Denoising). It, on the other hand, ignores the regression problem and does not assess the robustness of different models, instead of analyzing multiple defenses with a single classifier. FGSM and BIM attacks are used by another study \cite{rathore2020untargeted} to undertake untargeted, targeted, and universal adversarial attacks on time series classification problems. It does not, however, make any adversarial defenses and focuses solely on the classification problem.

\cite{mode2020adversarial} presents the concept of adversarial attacks (FGSM and BIM) on various deep learning models (CNN, LSTM, and GRU) for multivariate linear regression. Although it introduces the concept of adversarial attacks on regression tasks, it does not take account into the defense against the attacks. The results showed that DL regression models are vulnerable to adversarial attacks. On LSTM, CNN, RNN, and Multi-Head Attention Network, \cite{wu2022small}  examines two different attacks: FGSM and Adversarial attack with importance measurement (AAIM) (MHANet). However, it exclusively concentrates on the regression problem and makes no adversarial defenses.

None of the studies take both regression and categorization into account. Also, none of them investigate how vulnerable various models are to attacks and how well they can recover using defense.

\section{Problem Statement}
Previous works focus only on the adversarial attacks only for both regression and the classification problems in the series data. We have implemented the adversarial attacks and defense mechanism against attacks on both the classification and regression problems in the time series data. Adversarial attacks refer to finding adversarial examples for well-trained models. For a classification task, we use $f(x;\theta) $ to denote a model that maps an input to a discrete label set with k classes. Given a perturbation size $\epsilon$, the adversary tries to find a perturbation $\delta$ that maximizes the loss. Thus, the adversarial counterpart $ x^{i}$ of $x$ can be expressed as: $x^i=x+\delta$. Some of the common attacks for generating adversarial attacks are FGSM, BIM, and PGD. 

We designed a defense against these attacks using adversarial training for adversarial robustness. In general, Adversarial robustness is the model's performance on test data with adversarial attacks. Similar attacks techniques can also be used for adversarial training by expanding it to the min-max optimization problem. Min-max optimization problem consists of two parts: inner maximization which can be single or multiple steps (depending on the attack) is used to generate maximum perturbation and outer minimization is a single step stochastic gradient descent to minimize the loss of a model. 

\section{Methodology}

\begin{figure}[h]
    \centering
    \includegraphics[width=4.7in]{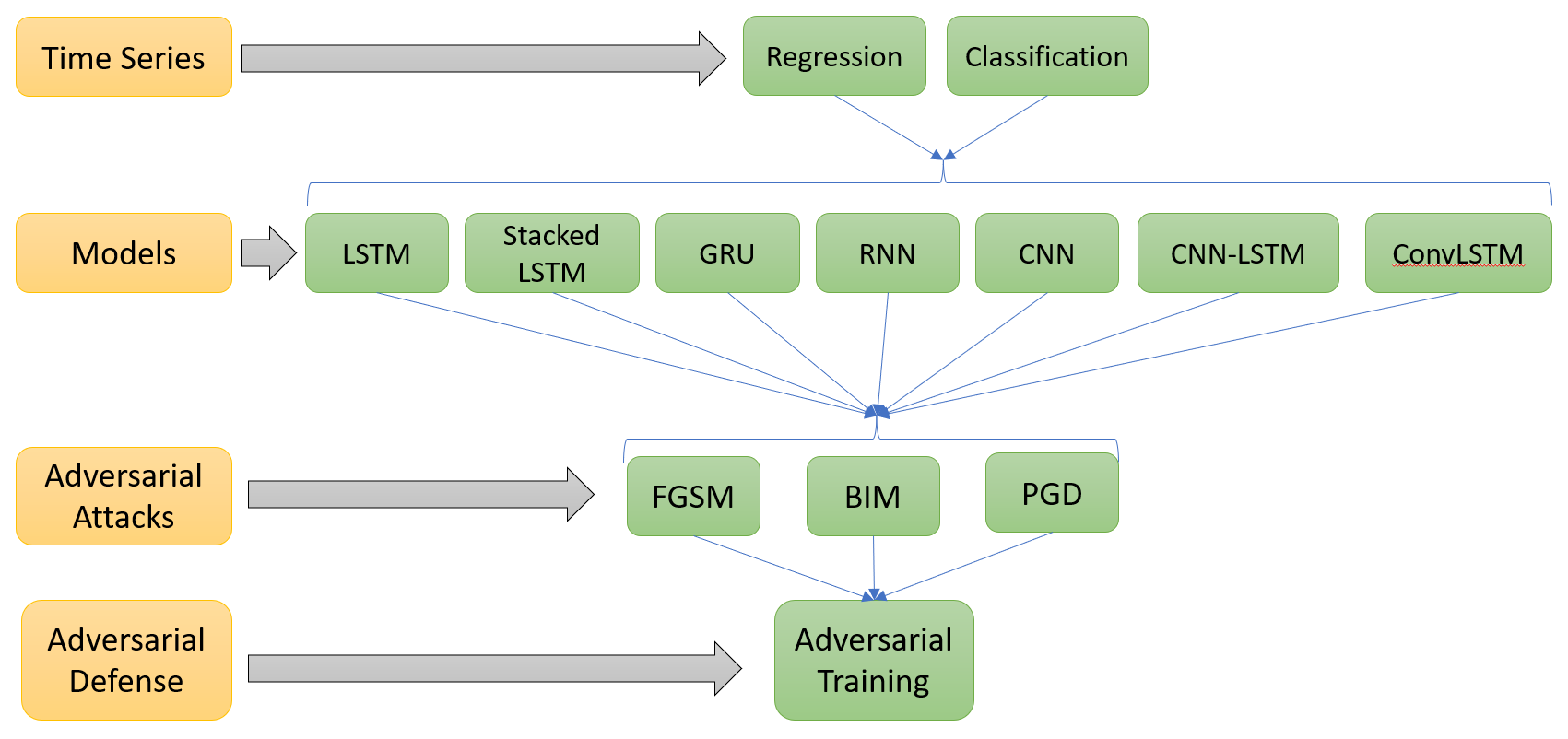}
    \caption{Overview of this study}
    \label{fig:overview}
\end{figure}

Various DL algorithms are trained for regression and classification problems for time series analysis. These models are attacked by generating perturbations on the test data. We implemented adversarial training as a defense mechanism to tackle the attacks. The subsections below introduce the models used for training, methods for generating adversarial attacks, and adversarial training mechanisms. Figure \ref{fig:overview} depicts the overview of this study.

\subsection{Models}
For the time series regression and classification, we used the following 7 models: LSTM (Long short-term memory), Stacked LSTM, GRU (Gated recurrent unit), RNN (Recurrent neural network), CNN (Convolutional neural network), CNN-LSTM, and ConvLSTM. The models are widely used for time series problems due to their capability of capturing sequential patterns. For sequential data, RNN, LSTM, and GRU are the go-to models. CNN is often used for time-series data as its convolutional capability is useful for modeling time-series patterns. CNN-LSTM and ConvLSTM combine the CNN and LSTM architectures, the latter of which is especially suitable for two-dimensional data. All the models are evaluated on time series classification and regression. Adversarial attacks and defense are employed in these models for analyzing their vulnerability and robustness. 

\subsection{Adversarial Attacks}

We have employed three well-known gradient-based attacks. Gradient-based attacks employ a perturbation for the input time series by altering the back-propagation process slightly. They use the gradients with respect to the input given the output for perturbation. All of these attacks are performed on the models for time series classification and regression. 

\subsubsection{Fast Sign Gradient Method (FGSM)} 
The fast gradient sign method (FGSM) \cite{goodfellow2014explaining} creates an adversarial example by leveraging the neural network's gradients. The approach creates a adversarial input that maximizes the loss for an original input by using the gradients of the loss with respect to the original input. The following equation can be used to summarize this: 
\begin{equation} \label{r-1}
    x_{adversarial} = x + \epsilon \times \textrm{sign}((\Delta_x J(\theta, x, y)) \\
\end{equation}
where $x_{adversarial}$ is the adversarial input, $x$ is the original input, $y$ is the original target variable, $\epsilon$ is the multiplier to ensure the perturbations are small, $\theta$ is the model parameters, and $J$ is the loss.

\subsubsection{Basic Iterative Method (BIM)}

Basic Iterative Method (BIM) \cite{kurakin2016adversarial} is an enhancement of the FGSM. It proposes repeating the FGSM step with a small step size ($\alpha$) and clipping the intermediate results after each step to verify that they are in the same neighborhood as the original input. It initializes the initial adversarial sample as the original input. Then it iteratively updates the adversarial sample while keeping it within the neighborhood of the original input. The size of the neighborhood is set by the hyperparameter $\epsilon$.  

\subsubsection{Projected Gradient Descent (PGD)}
Projected Gradient Descent (PGD) attack is quite similar to the BIM attack. The main difference is that PGD starts the adversarial example at a random location in the neighborhood (set by the multiplier $\epsilon$) and restarts it randomly, whereas BIM starts at the original input.

\subsection{Adversarial Training}

As a defense against these attacks, we performed adversarial training using similar attack methods for all the models in all the datasets. For each model, we created a new training set by combining the original training set and perturbed training sets using each attack. Thus for each dataset, we performed 21 adversarial pieces of training altogether 3 (FGSM, BIM, and PGD) for 7 different models. A new model with adversarial training is used to calculate adversarial robustness on the original perturbed testing data. 

Each adversarial training follows the principle of min-max optimization. Adversarial training with FGSM involves a single step of inner maximization using FGSM and a single step of outer minimization while adversarial training with BIM and PGD involves 5 steps of inner maximization using BIM and PGD and a single of outer minimization using stochastic gradient descent. 

\section{Experimental Evaluation}

\subsection{Data sets}
For classification, 1 data set is used as follows: 
\begin{itemize}
    \item Temperature\cite{muthukumar.j_2017}: It is derived from a Kaggle weather dataset. The data is based on
hourly temperature data for a city over ten years. We
use the first 23 time steps in the window as the
predictor variables and the last time step as the target variable.
\item Power Consumption\cite{powerdata}: It is derived from Kaggle which comes from PJM (regional transmission organization in the United States). It contains hourly energy consumption data in megawatts (MW) for the eastern part of the United State over three years. We
use the first 23 time steps in the window as the
predictor variables and the last time step as the target variable. 
\item Humidity\cite{muthukumar.j_2017}: It is derived from a Kaggle weather dataset. The data is based on
hourly humidity data for a city over ten years. We
use the first 23 time steps in the window as the
predictor variables and the last time step as the target variable.
\item Wind Speed\cite{muthukumar.j_2017}: It is derived from a Kaggle weather dataset. The data is based on
hourly wind speed data for a city over ten years. We
use the first 23 time steps in the window as the
predictor variables and the last time step as the target variable.
\item Hurricane\cite{landsea2013atlantic}: This corresponds to tropical cyclone intensity data obtained from the HURDAT2 database~\cite{landsea2013atlantic}. For each hurricane, wind speeds (intensities) were reported at every 6-hour interval. We use the first 23 time steps in the window as the predictor variables and the last time step as the target variable.
\item Eye State\cite{misc_eeg_eye_state_264}: This data is collected from UCI Machine Learning Repository-\cite{misc_eeg_eye_state_264}. It is from one continuous EEG measurement with the Emotiv EEG Neuroheadset. The duration of the measurement was 117 seconds. The eye state '1' indicates the eye-closed and '0' the eye-open state. 
\end{itemize}

\subsection{Experimental Setup}

We used different hyperparameters for the model's architecture and adversarial attacks. For the model's architecture, the number of layers, hidden nodes, and epochs is hyperparameters to be tuned. Similarly, for attacks the hyperparameters are perturbation budget $\epsilon$, step size $\alpha$, and the number of iterations in BIM and PGD. These hyperparameters are tuned differently depending on the attack and the model. We used $\epsilon$ between 0.1 to 0.3 and $\alpha$ between 0.1 and 0.2 depending on the model.

\subsection{Results}

We performed adversarial attacks and adversarial training on 5 regression problems (Power Consumption, Wind Speed, Temperature, Humidity, and Hurricane) and 1 classification problem (Eye State Classification) for time series analysis. Accuracy is used as a metric to determine the performance of the classification model. Table \ref{eye} depicts the results for our classification task. According to that, LSTM has the best performance for initial training with an accuracy of 82.46 \%. When three different adversarial attacks (FGSM, BIM, and PGD) are applied to the model, the accuracy drops to 58.54 \%, 72.83 \%, and 62.43\% respectively. The adversarial training increases the adversarial robustness of the model, boosting its accuracy to 81.28\%, 82.46\%, and 65.13\%.  

For the 5 regression tasks, RMSE is used as a metric for evaluating the performance of the model. For instance, the results of the Temperature dataset are summarized in Table \ref{temperature}, the GRU model during the initial training has the best performance with an RMSE of 0.01. When three different adversarial attacks (FGSM, BIM, and PGD) are applied to the model, the RMSE increases to 0.09, 0.09, and 0.07 respectively. The adversarial training increases the adversarial robustness of the model, reducing RMSE to 0.04,0.03, and 0.04 respectively. Other data sets show a similar pattern. Table \ref{power}, \ref{humidity}, \ref{wind}, \ref{hurricane} illustrate the rest of the four regression problems on Power Consumption, Wind Speed Humidity, and Hurricane dataset respectively. 

\begin{table}[H]
\centering
\begin{adjustbox}{width=0.75\textwidth}
\begin{tabular}{cc|ccccccc}
\multicolumn{2}{c|}{\textbf{Models}}                                                                                                  & LSTM & \begin{tabular}[c]{@{}c@{}}Stacked \\ LSTM\end{tabular} & GRU  & RNN  & CNN  & \begin{tabular}[c]{@{}c@{}}CNN-\\ LSTM\end{tabular} & ConvLSTM \\ \hline
\multicolumn{2}{c|}{\textbf{\begin{tabular}[c]{@{}c@{}}Accuracy for\\ models\end{tabular}}}                                           & \textbf{0.82} & \textbf{0.82}                                                    & 0.81 & 0.75 & 0.77 & 0.76                                                & 0.79     \\ \hline
\multicolumn{1}{c|}{\multirow{3}{*}{\textbf{\begin{tabular}[c]{@{}c@{}}Accuracy for\\ Adversarial\\ Attacks\end{tabular}}}}    & FGSM & \textbf{0.59} & 0.62                                                    & 0.59 & 0.65 & 0.66 & 0.54                                                & 0.61     \\
\multicolumn{1}{c|}{}                                                                                                          & BIM  & 0.72 & 0.74                                                    & 0.74 & 0.73 & 0.74 & 0.69                                                & 0.75     \\
\multicolumn{1}{c|}{}                                                                                                          & PGD  & 0.62 & 0.67                                                    & 0.61 & 0.66 & 0.65 & 0.59                                                & 0.56     \\ \hline
\multicolumn{1}{c|}{\multirow{3}{*}{\textbf{\begin{tabular}[c]{@{}c@{}}Accuracy after\\ Adversarial\\ Training\end{tabular}}}} & FGSM & \textbf{0.81} & 0.82                                                    & \textbf{0.81} & 0.71 & 0.78 & 0.73                                                & 0.81     \\
\multicolumn{1}{c|}{}                                                                                                          & BIM  & 0.82 & 0.82                                                    & 0.81 & 0.66 & 0.77 & 0.79                                                & 0.81     \\
\multicolumn{1}{c|}{}                                                                                                          & PGD  & 0.65 & 0.70                                                    & 0.65 & 0.63 & 0.67 & 0.68                                                & 0.68    
\end{tabular}
\end{adjustbox}
\vspace{0.3cm}
\caption{Adversarial attacks and defenses on different models for time series classification (Eye State)}
\label{eye}
\end{table}

\begin{table}[H]
\centering
\begin{adjustbox}{width=0.75\textwidth}
\begin{tabular}{cc|ccccccc}
\multicolumn{2}{c|}{\textbf{Models}}                                                                                              & LSTM & \begin{tabular}[c]{@{}c@{}}Stacked \\ LSTM\end{tabular} & GRU  & RNN  & CNN  & \begin{tabular}[c]{@{}c@{}}CNN-\\ LSTM\end{tabular} & ConvLSTM \\ \hline
\multicolumn{2}{c|}{\textbf{\begin{tabular}[c]{@{}c@{}}RMSE for \\ models\end{tabular}}}                                         & 0.03 & 0.03                                                    & \textbf{0.01} & 0.02 & 0.02 & 0.04                                                & 0.02     \\ \hline
\multicolumn{1}{c|}{\multirow{3}{*}{\textbf{\begin{tabular}[c]{@{}c@{}}RMSE for \\ Adversarial\\ Attacks\end{tabular}}}}  & FGSM & 0.09 & 0.1                                                     & \textbf{0.09} & 0.09 & 0.09 & 0.07                                                & 0.09     \\
\multicolumn{1}{c|}{}                                                                                                      & BIM  & 0.09 & 0.1                                                     & 0.09 & 0.09 & 0.09 & 0.07                                                & 0.09     \\
\multicolumn{1}{c|}{}                                                                                                      & PGD  & 0.06 & 0.08                                                    & 0.07 & 0.07 & 0.06 & 0.06                                                & 0.07     \\ \hline
\multicolumn{1}{c|}{\multirow{3}{*}{\textbf{\begin{tabular}[c]{@{}c@{}}RMSE after\\ Adversarial\\ Training\end{tabular}}}} & FGSM &  0.03  & 0.04                                                    & \textbf{0.04} & 0.04 & 0.03 & 0.04                                                & 0.03     \\
\multicolumn{1}{c|}{}                                                                                                      & BIM  & 0.06 & 0.06                                                    & 0.03 & 0.05 & 0.05 & 0.06                                                & 0.05     \\
\multicolumn{1}{c|}{}                                                                                                      & PGD  & 0.04 & 0.04                                                    & 0.04 & 0.04 & 0.05 & 0.05                                                & 0.04    
\end{tabular}
\end{adjustbox}
\vspace{0.3cm}
\caption{Adversarial attacks and defenses on different models for time series regression (Temperature)}
\label{temperature}
\end{table}

\begin{table}[H]
\centering
\begin{adjustbox}{width=0.75\textwidth}
\begin{tabular}{cc|ccccccc}
\multicolumn{2}{c|}{\textbf{Models}}                                                                                              & LSTM & \begin{tabular}[c]{@{}c@{}}Stacked \\ LSTM\end{tabular} & GRU  & RNN  & CNN  & \begin{tabular}[c]{@{}c@{}}CNN-\\ LSTM\end{tabular} & ConvLSTM \\ \hline
\multicolumn{2}{c|}{\textbf{\begin{tabular}[c]{@{}c@{}}RMSE for\\ models\end{tabular}}}                                           & 0.03 & 0.03                                                    & 0.02 & \textbf{0.02} & 0.02 & 0.03                                                & 0.02     \\ \hline   
\multicolumn{1}{c|}{\multirow{3}{*}{\textbf{\begin{tabular}[c]{@{}c@{}}RMSE for\\ Adversarial\\ Attacks\end{tabular}}}}    & FGSM & 0.26 & 0.22                                                    & 0.28 & \textbf{0.27} & 0.24 & 0.24                                                & 0.28     \\
\multicolumn{1}{c|}{}                                                                                                      & BIM  & 0.09 & 0.08                                                    & 0.09 & 0.09 & 0.1  & 0.1                                                 & 0.1      \\
\multicolumn{1}{c|}{}                                                                                                      & PGD  & 0.18 & 0.14                                                    & 0.25 & 0.22 & 0.21 & 0.22                                                & 0.18     \\ \hline 
\multicolumn{1}{c|}{\multirow{3}{*}{\textbf{\begin{tabular}[c]{@{}c@{}}RMSE after\\ Adversarial\\ Training\end{tabular}}}} & FGSM & 0.06 & 0.06                                                    & \textbf{0.06} & 0.07 & 0.04 & 0.05                                                & \textbf{0.05}     \\
\multicolumn{1}{c|}{}                                                                                                      & BIM  & 0.04 & 0.04                                                    & 0.05 & 0.05 & 0.03 & 0.04                                                & 0.04     \\
\multicolumn{1}{c|}{}                                                                                                      & PGD  & 0.08 & 0.07                                                    & 0.09 & 0.08 & 0.07 & 0.07                                                & 0.07    
\end{tabular}
\end{adjustbox}
\vspace{0.3cm}
\caption{Adversarial attacks and defenses on different models for time series regression (Power Consumption)}
\label{power}
\end{table}

\begin{table}[H]
\centering
\begin{adjustbox}{width=0.75\textwidth}
\begin{tabular}{cc|ccccccc}
\multicolumn{2}{c|}{\textbf{Models}}                                                                                              & LSTM & \begin{tabular}[c]{@{}c@{}}Stacked \\ LSTM\end{tabular} & GRU  & RNN  & CNN  & \begin{tabular}[c]{@{}c@{}}CNN-\\ LSTM\end{tabular} & ConvLSTM \\ \hline
\multicolumn{2}{c|}{\textbf{\begin{tabular}[c]{@{}c@{}}RMSE for\\ models\end{tabular}}}                                           & 0.04 & 0.04                                                    & 0.04 & \textbf{0.03} & 0.04 & 0.05                                                & 0.03     \\ \hline
\multicolumn{1}{c|}{\multirow{3}{*}{\textbf{\begin{tabular}[c]{@{}c@{}}RMSE for\\ Adversarial\\ Attacks\end{tabular}}}}    & FGSM & 0.13 & 0.12                                                    & 0.16 & \textbf{0.16} & 0.14 & 0.18                                                & 0.14     \\
\multicolumn{1}{c|}{}                                                                                                      & BIM  & 0.13 & 0.13                                                    & 0.16 & 0.16 & 0.14 & 0.18                                                & 0.14     \\
\multicolumn{1}{c|}{}                                                                                                      & PGD  & 0.12 & 0.1                                                     & 0.14 & 0.13 & 0.12 & 0.16                                                & 0.13     \\ \hline
\multicolumn{1}{c|}{\multirow{3}{*}{\textbf{\begin{tabular}[c]{@{}c@{}}RMSE after\\ Adversarial\\ Training\end{tabular}}}} & FGSM & 0.08 & 0.08                                                    & \textbf{0.1}  & \textbf{0.1}  & 0.08 & 0.14                                                & 0.08      \\
\multicolumn{1}{c|}{}                                                                                                      & BIM  & 0.11 & 0.1                                                     & 0.12 & 0.13 & 0.11 & 0.14                                                & 0.13     \\
\multicolumn{1}{c|}{}                                                                                                      & PGD  & 0.08 & 0.09                                                    & 0.1  & 0.1  & 0.1  & 0.13                                                & 0.1     
\end{tabular}
\end{adjustbox}
\vspace{0.3cm}
\caption{Adversarial attacks and defenses on different models for time series regression (Humidity)}
\label{humidity}
\end{table}

\begin{table}[H]
\centering
\begin{adjustbox}{width=0.75\textwidth}
\begin{tabular}{cc|ccccccc}
\multicolumn{2}{c|}{\textbf{Models}}                                                                                              & LSTM & \begin{tabular}[c]{@{}c@{}}Stacked \\ LSTM\end{tabular} & GRU  & RNN  & CNN  & \begin{tabular}[c]{@{}c@{}}CNN-\\ LSTM\end{tabular} & ConvLSTM \\ \hline
\multicolumn{2}{c|}{\textbf{\begin{tabular}[c]{@{}c@{}}RMSE for\\ models\end{tabular}}}                                           & 0.04 & 0.04                                                    & \textbf{0.01} & 0.02 & 0.02 & 0.04                                                & 0.04     \\ \hline
\multicolumn{1}{c|}{\multirow{3}{*}{\textbf{\begin{tabular}[c]{@{}c@{}}RMSE for\\ Adversarial\\ Attacks\end{tabular}}}}    & FGSM & 0.12 & 0.12                                                    & \textbf{0.09} & 0.09 & 0.09 & 0.07                                                & 0.11     \\
\multicolumn{1}{c|}{}                                                                                                      & BIM  & 0.12 & 0.12                                                    & 0.09 & 0.09 & 0.09 & 0.07                                                & 0.11     \\
\multicolumn{1}{c|}{}                                                                                                      & PGD  & 0.09 & 0.09                                                    & 0.07 & 0.07 & 0.06 & 0.06                                                & 0.11     \\ \hline
\multicolumn{1}{c|}{\multirow{3}{*}{\textbf{\begin{tabular}[c]{@{}c@{}}RMSE after\\ Adversarial\\ Training\end{tabular}}}} & FGSM & 0.06 & 0.07                                                    & \textbf{0.04} & 0.04 & 0.05 & 0.04                                                & 0.08     \\
\multicolumn{1}{c|}{}                                                                                                      & BIM  & 0.09 & 0.1                                                     & 0.06 & 0.05 & 0.05 & 0.04                                                & 0.1      \\
\multicolumn{1}{c|}{}                                                                                                      & PGD  & 0.06 & 0.06                                                    & 0.04 & 0.04 & 0.05 & 0.05                                                & 0.08    
\end{tabular}
\end{adjustbox}
\vspace{0.3cm}
\caption{Adversarial attacks and defenses on different models for time series regression (Wind Speed)}
\label{wind}
\end{table}

\begin{table}[h]
\centering
\begin{adjustbox}{width=0.8\textwidth}
\begin{tabular}{cc|ccccccc}
\multicolumn{2}{c|}{\textbf{Models}}                                                                                              & LSTM & \begin{tabular}[c]{@{}c@{}}Stacked \\ LSTM\end{tabular} & GRU  & RNN  & CNN  & \begin{tabular}[c]{@{}c@{}}CNN-\\ LSTM\end{tabular} & ConvLSTM \\ \hline
\multicolumn{2}{c|}{\textbf{\begin{tabular}[c]{@{}c@{}}RMSE for\\ models\end{tabular}}}                                           & 0.21 & 0.22                                                    & \textbf{0.19} & 0.19 & 0.21 & 0.43                                                & 0.19     \\ \hline
\multicolumn{1}{c|}{\multirow{3}{*}{\textbf{\begin{tabular}[c]{@{}c@{}}RMSE for\\ Adversarial\\ Attacks\end{tabular}}}}    & FGSM & 0.3  & 0.32                                                    & 0\textbf{0.39} & 0.38 & 0.39 & 0.6                                                 & 0.36     \\
\multicolumn{1}{c|}{}                                                                                                      & BIM  & 0.27 & 0.29                                                    & 0.35 & 0.34 & 0.35 & 0.56                                                & 0.32     \\
\multicolumn{1}{c|}{}                                                                                                      & PGD  & 0.28 & 0.29                                                    & 0.36 & 0.35 & 0.36 & 0.55                                                & 0.34     \\ \hline
\multicolumn{1}{c|}{\multirow{3}{*}{\textbf{\begin{tabular}[c]{@{}c@{}}RMSE after\\ Adversarial\\ Training\end{tabular}}}} & FGSM & 0.29 & 0.28                                                    & \textbf{0.29} & 0.29 & 0.29 & 0.55                                                & 0.3      \\
\multicolumn{1}{c|}{}                                                                                                      & BIM  & 0.26 & 0.29                                                    & 0.23 & 0.27 & 0.3  & 0.57                                                & 0.27     \\
\multicolumn{1}{c|}{}                                                                                                      & PGD  & 0.3  & 0.28                                                    & 0.26 & 0.28 & 0.29 & 0.51                                                & 0.3     
\end{tabular}
\end{adjustbox}
\vspace{0.3cm}
\caption{Adversarial attacks and defenses on different models for time series regression (Hurricane Intensity)}
\label{hurricane}
\end{table}

\begin{figure}[h]
    \centering
    \includegraphics[width=3.5in]{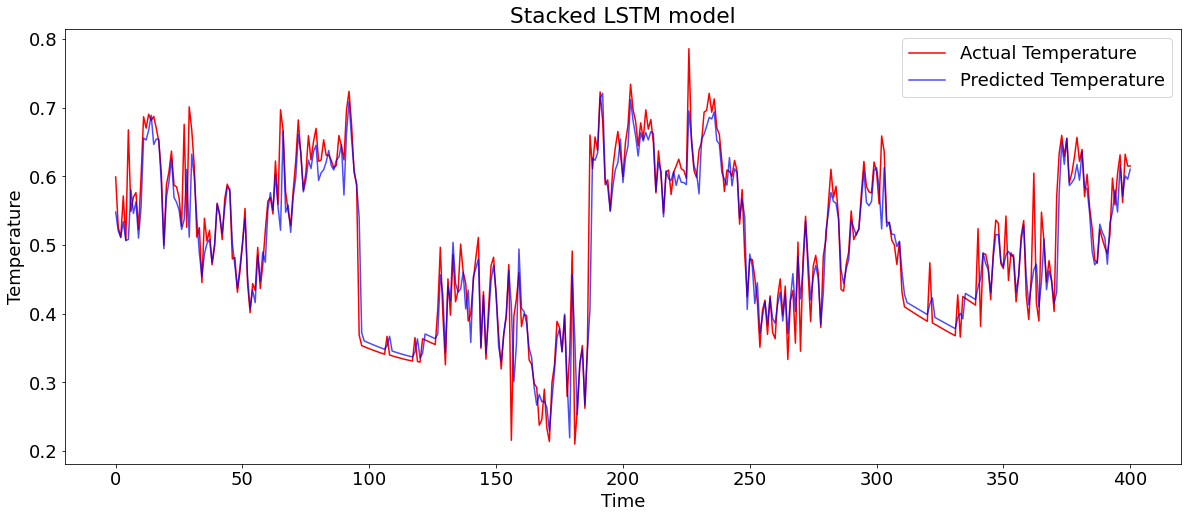}
    \caption{Actual Temperature vs Predicted Temperature by LSTM model for the Temperature dataset}
    \label{fig:2}
\end{figure}
 \begin{figure}[H]
    \centering
    \subfloat[1]{{\includegraphics[width=0.45\textwidth]{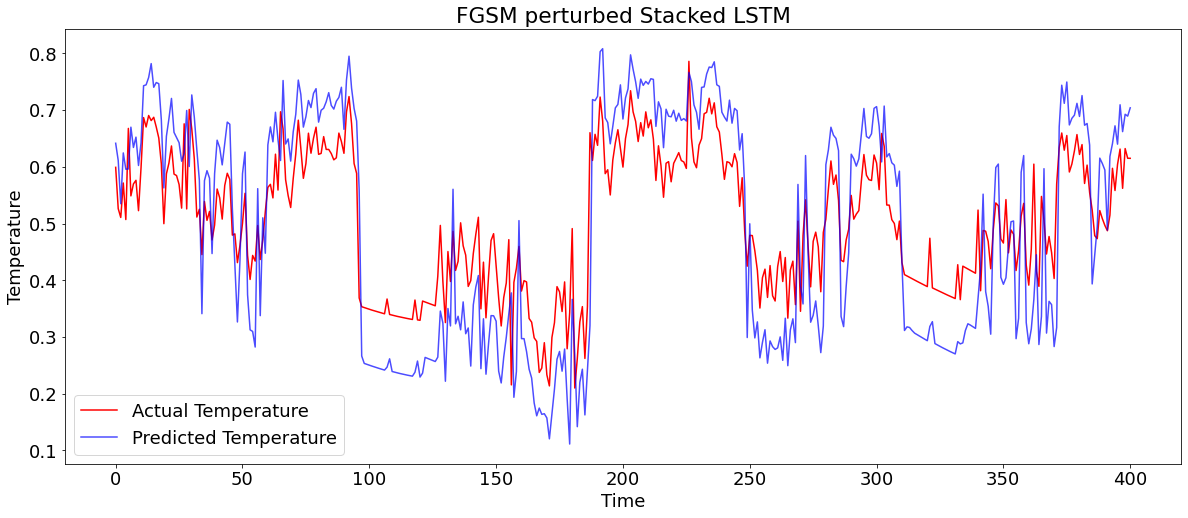}}}
    \hfill
    \subfloat[2]{{\includegraphics[width=0.45\textwidth]{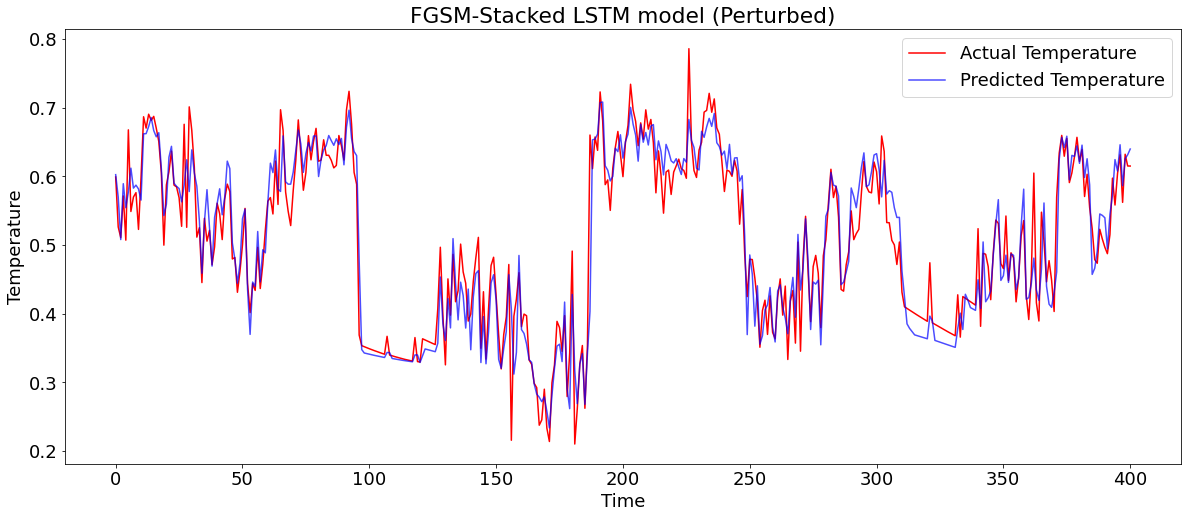}}}
   
    \subfloat[3]{{\includegraphics[width=0.45\textwidth]{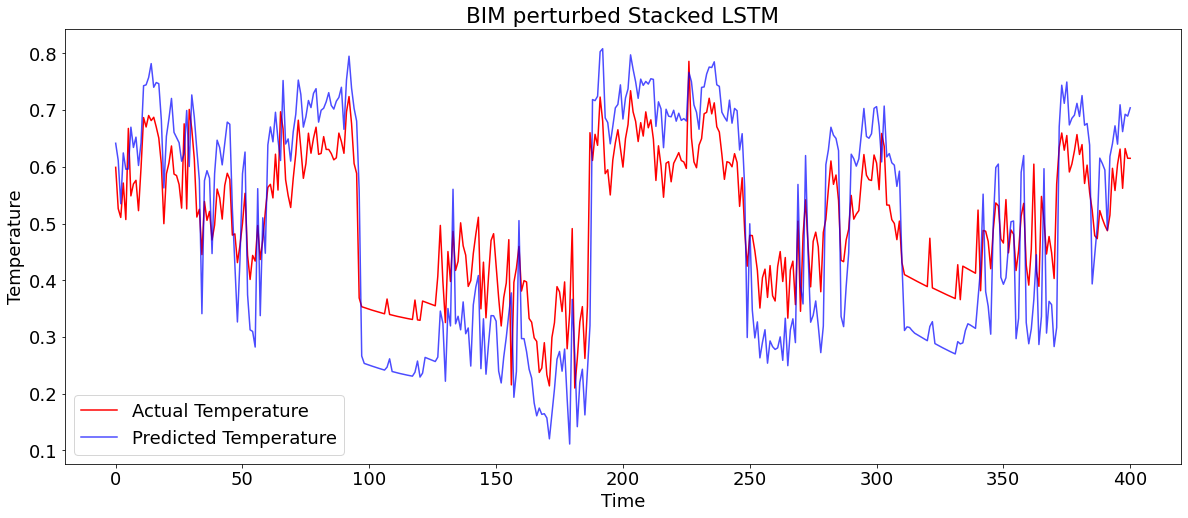}}}
    \hfill
    \subfloat[4]{{\includegraphics[width=0.45\textwidth]{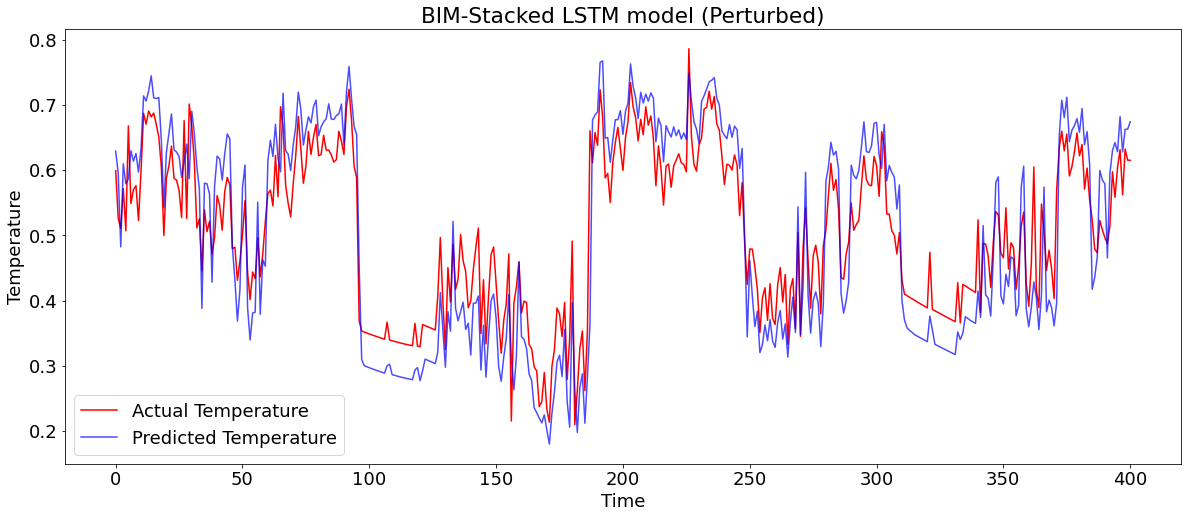}}}
   
    \subfloat[5]{{\includegraphics[width=0.45\textwidth]{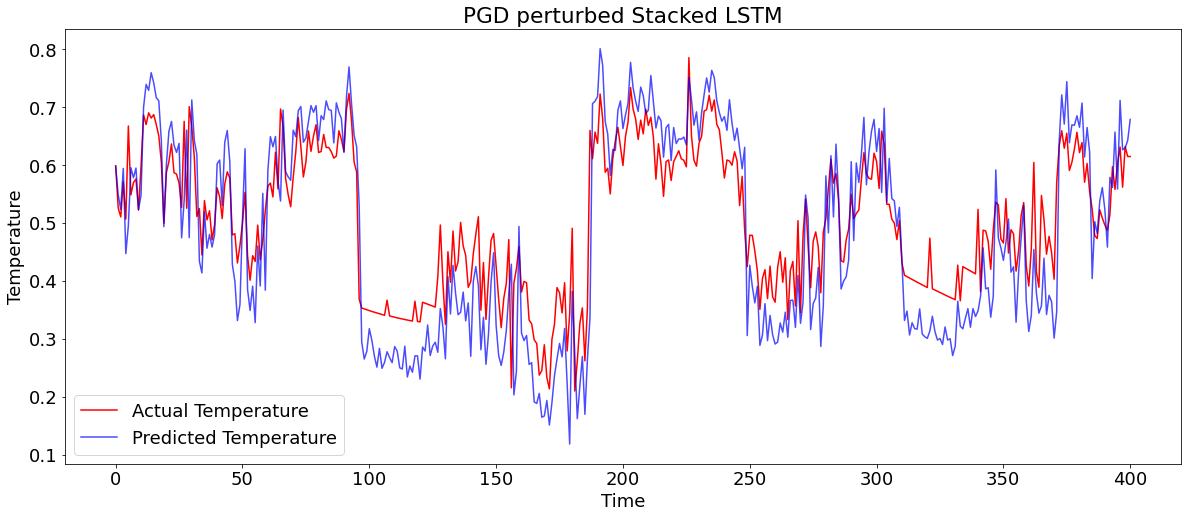}}}
    \hfill
    \subfloat[6]{{\includegraphics[width=0.45\textwidth]{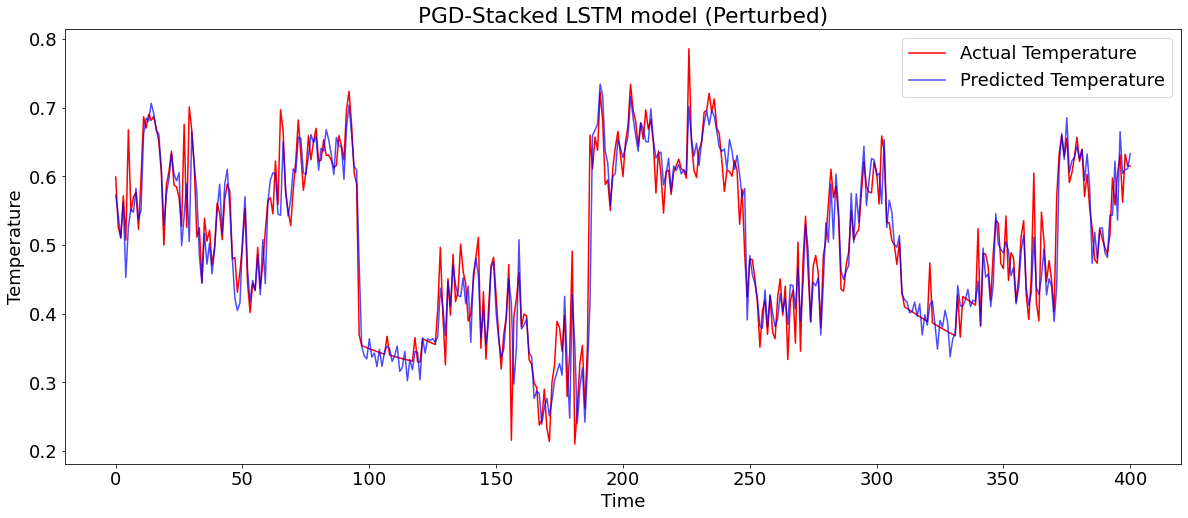}}}

    \caption{Time series plot for actual temperature vs predicted temperature in LSTM models after adversarial attacks and adversarial training( [1]Adv attack with FGSM [2] Adv training with FGSM [3] Adv attack with BIM [4] Adv training with BIM [5] Adv attack with PGD [6] Adv training with PGD)}%
\end{figure}

\section{Insights/Discussion}
Based on our experiments on the data sets we used, we observed that FGSM generates a much stronger attack than both BIM and PGD. BIM performs poorly among these three attacks. BIM's initial perturbation is set to the original input. It might be the reason that it can not perturb much as it starts from the original input. Apart from attack generation, FGSM also has higher training efficiency than both BIM and PGD. 

For the adversarial defense, recovering from the FGSM perturbed model is the easiest one. But, recovering from a PGD attack is the hardest. PGD's random initialization and random restarts might make it harder for recovery.  

Among different time series models, all-time series models are vulnerable to adversarial attacks. GRU and RNN seem to be more vulnerable than the others. In terms of recovery, all models can recover from the attacks, but LSTM and GRU can recover slightly better.

\section{Individual Contributions}
\textbf{Bidhan}:Literature review of \cite{fawaz2019adversarial}\cite{harford2020adversarial} \cite{mode2020adversarial}; implemented 3 models (LSTM, RNN and Stacked-LSTM); Implemented Adversarial training part.

\textbf{Asadullah Hill Galib}:Literature review of \cite{siddiqui2020benchmarking}\cite{wu2022small} \cite{rathore2020untargeted}; implemented 4 models (GRU, CNN, CNN-LSTM, ConvLSTM); Implemented Adversarial attacks part.

\section{Conclusion}

In this paper, we look at how vulnerable different time series models are under adversarial attacks and how well they recover using adversarial defense. We evaluated 7-time series models on three gradient-based attack mechanisms and adversarial training-based defense against them for both regression and classification problems. FGSM attack has a much stronger effect on generating adversarial attacks and efficient adversarial training. All models, especially GRU and RNN, appear to be vulnerable, according to the findings. In addition, LSTM and GRU have higher defense recovery. In terms of attacks, FGSM outperforms the competition. It's more difficult to recover from a PGD attack than it is for other types of attacks. In the future, we will extend this work by incorporating more types of attacks and defense mechanisms. 


\bibliographystyle{unsrt}

\end{document}